\documentclass{article}

\usepackage{arxiv}

\usepackage[utf8]{inputenc} 
\usepackage[T1]{fontenc}    
\usepackage{hyperref}       
\usepackage{url}            
\usepackage{booktabs}       
\usepackage{amsfonts}       
\usepackage{nicefrac}       
\usepackage{microtype}      
\usepackage{lipsum}		
\usepackage{graphicx}
\usepackage[numbers]{natbib}
\usepackage{doi}

\title{Discovering Behavioral Predispositions in Data to Improve Human Activity Recognition}

\author{Maximilian Popko\\
University of Rostock, Germany\\
maximilian.popko@uni-rostock.de
\And
Sebastian Bader\\
University of Rostock, Germany\\
sebastian.bader@uni-rostock.de
\And
Stefan L\"udtke\\
University of Mannheim, Germany\\
luedtke@es.uni-mannheim.de\\
\And
Thomas Kirste\\
University of Rostock, Germany\\
thomas.kirste@uni-rostock.de}

\renewcommand{\shorttitle}{Discovering Behavioral Predispositions in Data to Improve Human Activity Recognition}

\hypersetup{
pdftitle={Discovering Behavioral Predispositions in Data to Improve Human Activity Recognition},
pdfauthor={Maximilian Popko, Sebastian Bader, Stefan Lüdtke, Thomas Kirste},
pdfkeywords={Human activity recognition, Wearable sensors, Machine learning, Clustering},
}

\begin{document}
\maketitle

\begin{abstract}
The automatic, sensor-based assessment of challenging behavior of persons with dementia is an important task to support the selection of interventions.  However, predicting behaviors like apathy and agitation is challenging due to the large inter- and intra-patient variability. Goal of this paper is to improve the recognition performance by making use of the observation that patients tend to show specific behaviors at certain times of the day or week. We propose to identify such segments of similar behavior via clustering the distributions of annotations of the time segments. All time segments within a cluster then consist of similar behaviors and thus indicate a behavioral predisposition (BPD). We utilize BPDs by training a classifier for each BPD. Empirically, we demonstrate that when the BPD per time segment is known, activity recognition performance can be substantially improved.
\end{abstract}

\keywords{Human activity recognition \and Wearable sensors \and Machine learning \and Clustering}

\section{Introduction}
The recognition of human behavior is important for many domains, like sports, the analysis of manual work processes, or as a component for situation-aware assistants. As another example, sensor-based Human Activity Recognition (HAR) can also be used for the automatic assessment of behavioral symptoms of people with dementia, like apathy or agitation. In contrast to questionnaires like the Cohen-Mansfield Agitation Inventory \citep{cmai}, HAR allows the real-time, objective assessment of symptoms, supporting caregivers in selecting appropriate interventions. 
However, accurate HAR is challenging due to the high inter- and intra-subject variability of movement. Specifically, the automatic recognition of challenging behavior from wearable sensor data has been of limited success so far \citep{harchall2}.

Multiple approaches have been proposed to improve HAR accuracy, by making use of additional external knowledge \citep{leehar, xie, yordanovacooking, rueda, luedtkenn, inoue}. 
In this paper, we investigate how HAR performance can be improved by making use of the fact that there is a tendency of specific subjects to show specific behaviors at certain times of the day or week. For example, sundowning is the increase in restlessness of some people with dementia in the evening \citep{sundowning}.
The knowledge of such behavioral predispositions (BPDs) could serve as a useful prior to increase HAR accuracy. 

\begin{figure*}
  \centering
  \includegraphics[width=\linewidth]{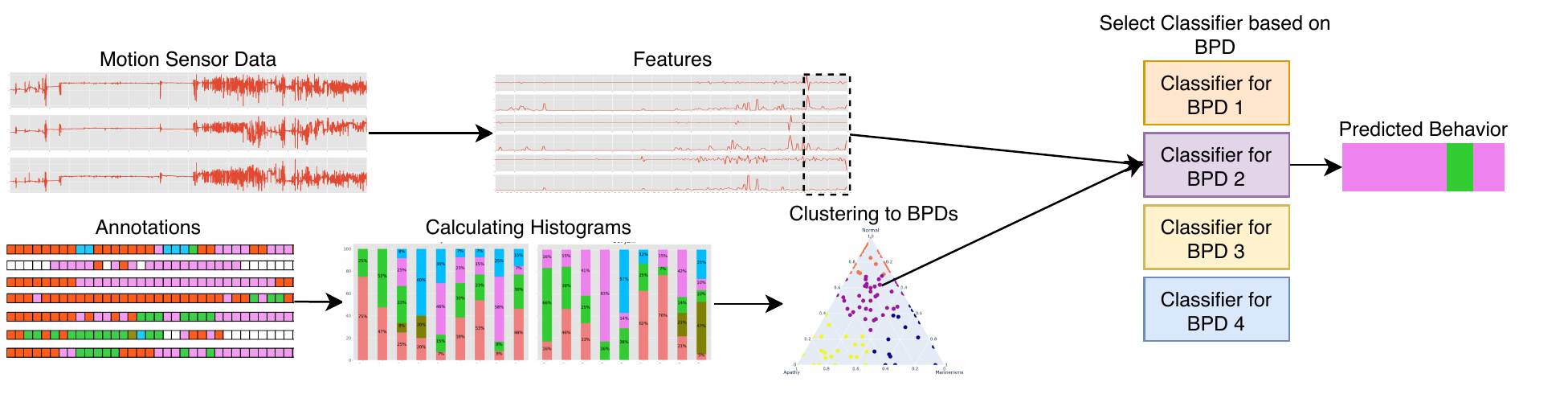}
  \caption{Overview of the concept. We calculate the distribution of annotations for a given segment of the day (histograms). We then cluster these distributions. Next, we use the cluster label (BPD) to select the classifier (color indicates the corresponding classifier). The classifier predicts the behavior from the features of the motion sensor data.}
  \label{fig:overview}
\end{figure*}

Previous work relied on external knowledge to define BPDs. For example, L\"udtke et al. \citep{luedtkenn} partitioned warehouse order picking into distinct process steps, the knowledge of which can inform HAR. However, in the context of dementia care, there are no established BPDs that are externally given. 
The contribution of this paper is to identify such BPDs in a data-driven way:  We calculate the distribution of annotations in time segments of a day, and then cluster these histograms. This results in clusters of time frames in which the subjects behaved similarly. Each cluster forms a BPD. Given the BPD, we then train a classifier specifically for data recorded within the BPD, i.e., there is one classifier for each cluster. This classifier is thus biased towards the more frequent classes in the BPD. An overview of this process can be found in figure \ref{fig:overview}.  In summary, this paper provides the following contributions:
\begin{itemize}
\item A clustering-based method for the data-driven construction of behavioral predispositions (BPDs)
\item An approach for utilizing these BPDs in HAR
\item An evaluation of this approach for the sensor-based recognition of behavioral symptom of persons with dementia
\end{itemize}
Empirically, our approach leads to a substantial increase of HAR performance, when assuming that the BPD for each time segment is given.

\section{Data and Observations}
This section presents the insideDem framework, as well as the data set. We make observations about the distribution of annotations given time frames, which will then lead to the construction of the BPDs and the utilization of them in the methodology section \ref{sec:methods}.

\subsection{insideDem Project}
Part of the insideDem framework was the recording of the behaviours of dementia diagnozed patients \citep{teipelinsidedem}. In two nursing homes, real-time observations and sensor data of a total of 17 residents (11 women and 6 men) with moderate to severe dementia were recorded. The behaviour of persons in one nursing home was also recorded on video. The age range was 73 to 94 years. All persons were under psycho-pharmacological treatment. Dementia medication was taken by 8 residents of the nursing homes. The behaviour of the persons was recorded in real time by an expert in the sitting room of the nursing home. This room was accessible to all residents and their visitors as well as the staff of the nursing home. The behaviour was annotated every five minutes following the annotation scheme which was developed for the insideDem project. This scheme is based on the neuropsychiatric inventory \citep{npi} and the Cohens-Mansfield agitation inventory \citep{cmai}. It consists of seven different behaviours: apathy, general restlessness, mannerisms, pacing, aggression, trying to get to a different place, normal behaviour. 

During the day, between 8 a.m. and 6 p.m., the subjects wore two wristbands, which were attached to the dominant hand and the ankle. During the night, a band was worn on the ankle. The sensors recorded accelerations and rotations as well as the sound level of the environment, the light conditions and the sound pressure. In this work, we only use the acceleration sensor data. 

\subsection{Observations}
\begin{figure}
  \centering
  \includegraphics[width=0.8\linewidth]{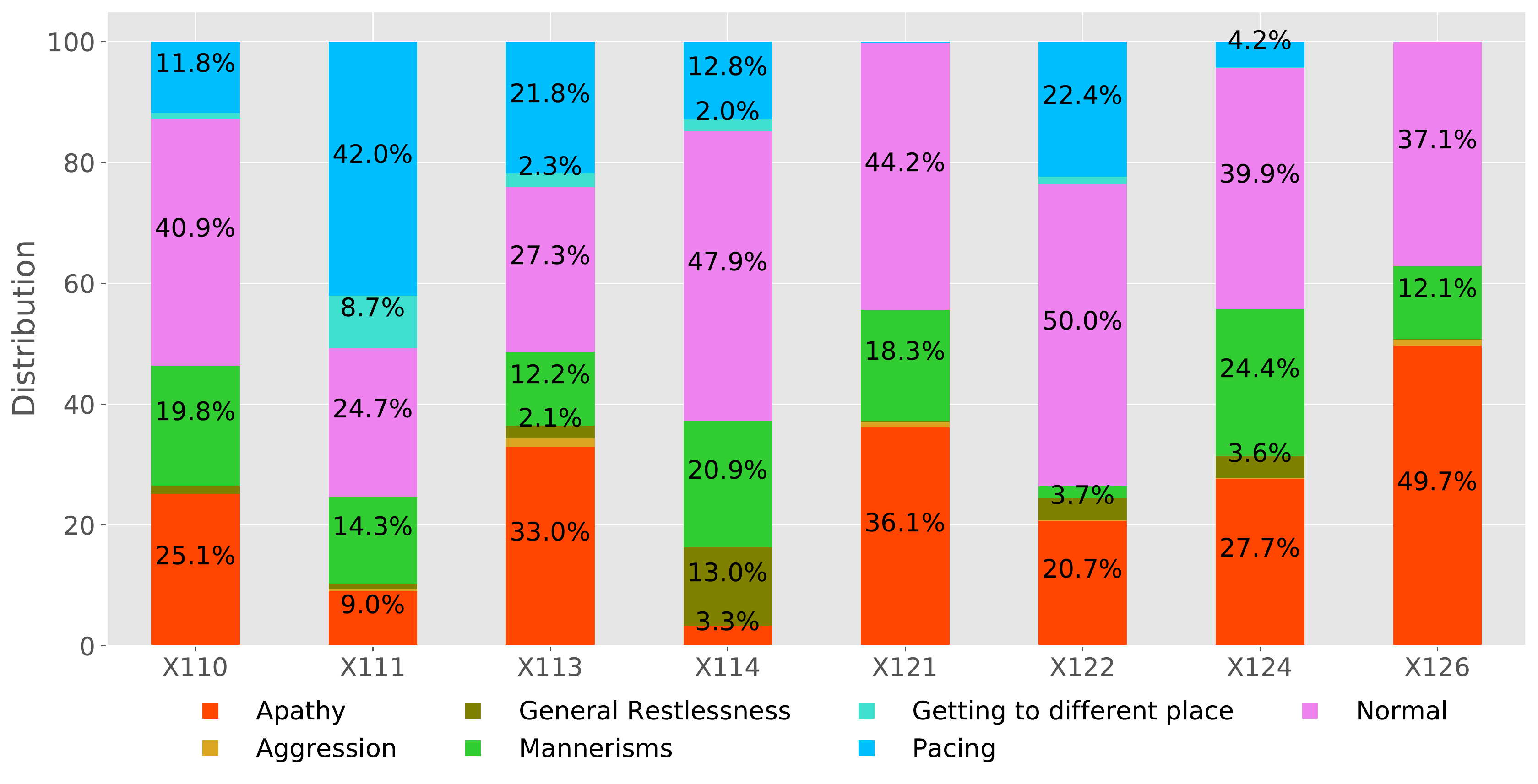}
  \caption{Distribution of annotations per person over all days of recordings.}
  \label{fig:hist}
\end{figure}

Figure \ref{fig:hist} shows the distribution of annotations for each subject. The behaviour of the eight patients were observed over the course of 29 days and accumulate around 17,000 annotated five-minute intervals. For every subject the the behaviours occur in different portions. As example, the patient X111 showed 9\% apathy where as X126 showed nearly 50\% apathy during the time span of 29 days. However, for  X111, pacing was mostly annotated but X126 nearly showed no pacing. 

In the literature events for triggering agitated behaviors were observed. The sunset phenomenon as described in \citep{sundowning} may be such a trigger for one BPD. At sunset people with dementia often show more agitated behavior than at other times of the day. Also, the fact of boredom can trigger agitated behavior. If there is nothing planned for an afternoon, dementia diagnosed persons might show agitated behavior because of the boredom \citep{trigger}. As a result, it is reasonable to look into noticeable time frames of the day which show one person behaving agitated more often.

\begin{figure}
  \centering
  \includegraphics[width=0.8\linewidth]{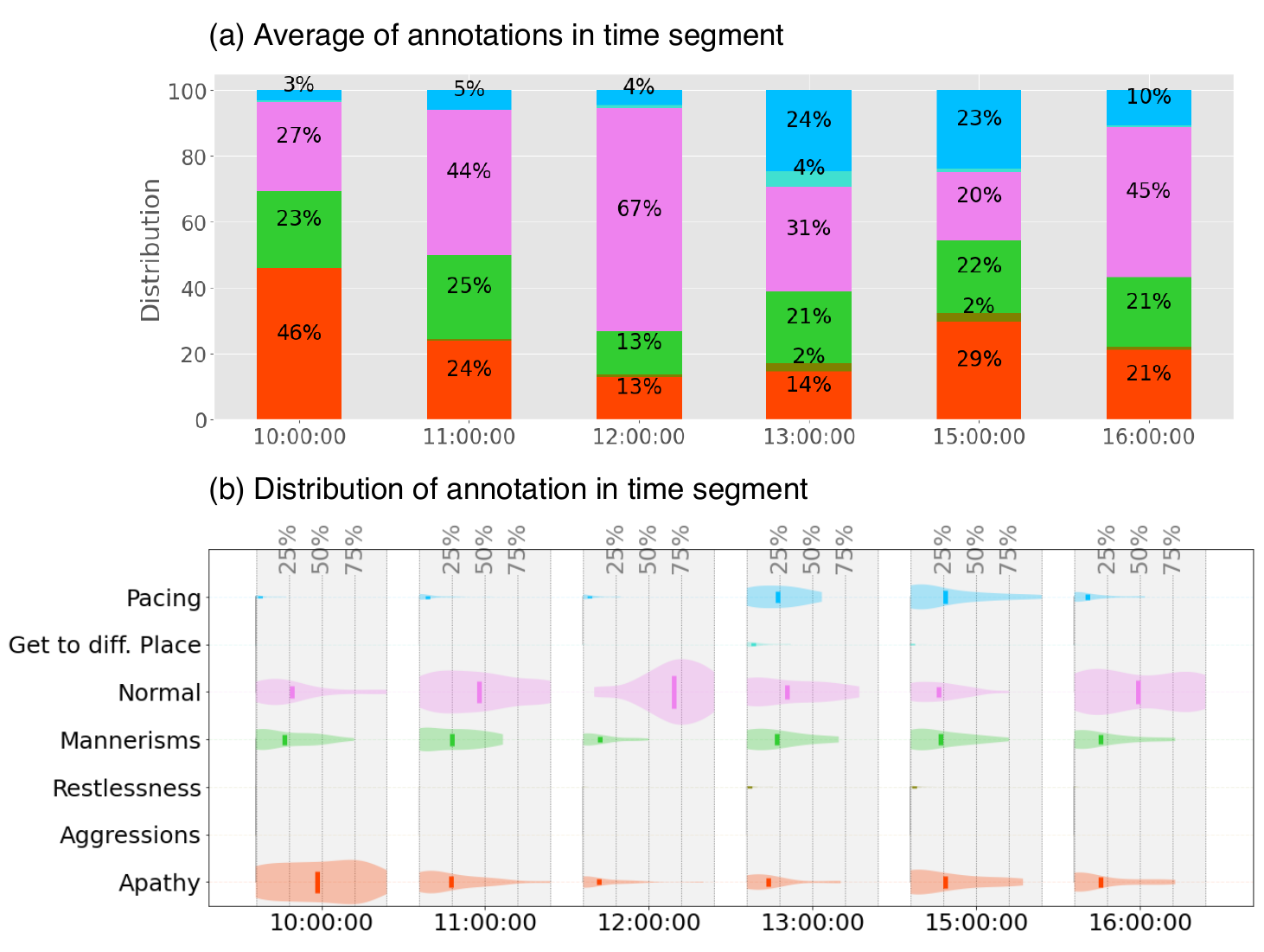}
  \caption{Percentual distribution of annotations for one subject and one-hour intervals of the day. (a) shows the average of the annotation and (b) describes the distribution (variance).}
  \label{fig:hist_violin}
\end{figure}

The figure \ref{fig:hist_violin} (a) now shows the average occurrences of the annotations for one-hour segments of the day. The represented subject behaves in the morning more apathetic than in the later course of the day. Also at noon, the patient shows more normal behavior. Pacing is mostly happening in the afternoon. Figure \ref{fig:hist_violin} (b) shows the variances of the annotations between each day. This plot represents the distribution of the proportion each behavior has on the time segment every day. Between 10 and 11 a.m. we saw the patient behaves apathetically to 46\% in plot (a). However, in plot (b) we see that the apathy part can differ from 0\% to 100\% with equal probability from day to day.

\section{Methods}
\label{sec:methods}
Behavioural predispositions (BPD) describe the tendency to show specific behaviours. In this section, we propose a method to form such BPDs from the annotations of behaviors. In the following, we present a simple model that can use these newly found BPDs to simplify the classification.

\subsection{Extracting behavioural predispositions}
BPD describe the urge to show specific kinds of behaviors. Knowing such would intuitively mean that the prediction of the current actual behavior becomes easier since it provides an informative prior on the behavior of the subject. Unfortunately an expert model describing BPD for patients with dementia does not exist and thus the aim is to derive them from the data of the insideDem project.

\subsubsection{Time-based BPD}

Following the observations of figure \ref{fig:hist_violin} (a), one can argue to use a similar approach to \citep{inoue} or define that every BPD is one segment of the day. This means from 10 to 11 a.m. the subject always behaves similarly and by knowing the current time, we can use this information to improve the classification. However, this approach has the drawback of high variances of the annotations between each day which lead to BPDs that do not carry a lot of information about the actual behavior.

\subsubsection{Clustering-based BPD}

To overcome the problem of high variances, we propose to find similar distributions globally over all days and time segments instead of grouping them by the time of the day. Let $p(A_t)$ be the probability distribution of the annotation for the day and time segment $t$. Note that $t$ describes a specific time frame e.g. 06--15 10:00 to 06--15 11:00. This categorical distribution can be obtained by counting the relative frequencies of the annotations in the segment. This counted distribution is called a histogram. The BPD represents the tendency to behave in a certain way. Given a set of histograms, the task is to divide this set into groups. These groups should contain similar histograms. With the help of the k-means clustering algorithm, clusters of histograms can be generated which are grouped according to their spatial similarity (Euclidean distance). Each of these clusters describes a behavioral disposition. The histograms of the cluster $D=d$ therefore represent the distribution $p(A \, | \, D = d)$. 

\subsection{Classification of Sensor Data}
In the scope of this work, we want to show what impact the complete knowledge of the clustered BPD can have on the classification. Therefore, we assume the BPD, i.e. the cluster, as given. Note that this would not be the case in a real world environment. Given a BPD $d$, we then have a prior knowledge of the annotation $p(A \, | \, d) $. To make use of this knowledge, we train a different classifier for each BPD. Thus, each classifier is trained with biased data and is therefore biased to more likely predict the classes that are most presently in this BPD. The classifier can be seen as expert for its BPD. The model $m$ maps sensor data $s$ and the BPD $d$ to an annotation $a$ by choosing the classifier $c_d$ that was specifically trained for $d$. The classifier then maps $s$ to $a$:
\begin{displaymath}
  m(s,d) = c_d(s) = a \, .
\end{displaymath}

\section{Experimental Evaluation}
\label{sec:eval}
The overall goal of the experiments are the evaluation of the impact of the  BPD on the HAR performance. In the following, we describe the experimental design in more detail.

\subsection{Experimental Design}
\begin{table*}
\centering
\caption{Factors and levels of the experimental design}
\begin{tabular}{p{2cm}p{5cm}p{6cm}}
\toprule
Factors & Levels & Description \\
\midrule
Clustering & K-means, time-based & Methods for generating the BPDs \\
\#{}-clusters & 1, \dots , 20 & Number of BPDs \\
Segment lengths & 0, \dots , 120 minutes (k-means)  & Length of the segment over which the histogram is calculated, only for k-means approach \\
Classifiers & Majority, naive Bayes, SVM & Classifiers to predict the annotation from sensor data \\
Subjects & X110, X111, X113, X114, X121, X122, X124, X126 & One model for each subject \\
\bottomrule
\end{tabular}
\label{tab:parameter}
\end{table*}
To evaluate the impact of the knowledge of the BPDs on HAR performance, we use a factorial design. The factors and levels are presented in table \ref{tab:parameter}. For each subject, a model with a different clustering method, number of clusters (i.e. BPDs), segment length, and classifier is trained. We randomly select 70\% of the data for training and 30\% for testing the model. Every combination of the levels is evaluated 10 times. Different clusters and segment lengths are used to test the expressiveness of the BPD. The segment length plays an important role for the k-means clustering strategy as the length is used to define the size over which the histogram is calculated and then clustered. At a segment length of 30 minutes, we obtain around 400 histograms per subject. We then cluster the histograms of a single subject, resulting in BPDs specifically for this subject. High number of clusters and small segment lengths yield finer BPDs. Note that one cluster means we have no further information about the BPD and thus gives us a baseline of performance. The time-based clustering method describes that a BPD only depends on the time. E.g., having 10 clusters and time-based clustering this method would split the day into 10 parts and for every part, one classifier is trained. We use three different methods for classifying the sensor data. The majority method functions as a baseline by always outputting the most frequent annotation of the BPD. The naive Bayes (NB) method shows the performance of a simplistic classifier. The NB uses for every BPD a different prior and observation model. Thirdly, a support vector machine (SVM) is trained. For preprocessing the accelerator data, features were calculated following the work of \citep{features}. The original data is sampled in 50 Hz. Feature were calculated over a window size of one minute and an overlap of 50\%. Statistical features, spatial relations between the three axis and frequency features were used. As measure of accuracy, we use $F_1$ score.

\subsection{Results and Discussion}

\begin{figure*}
  \centering
  \includegraphics[width=\linewidth]{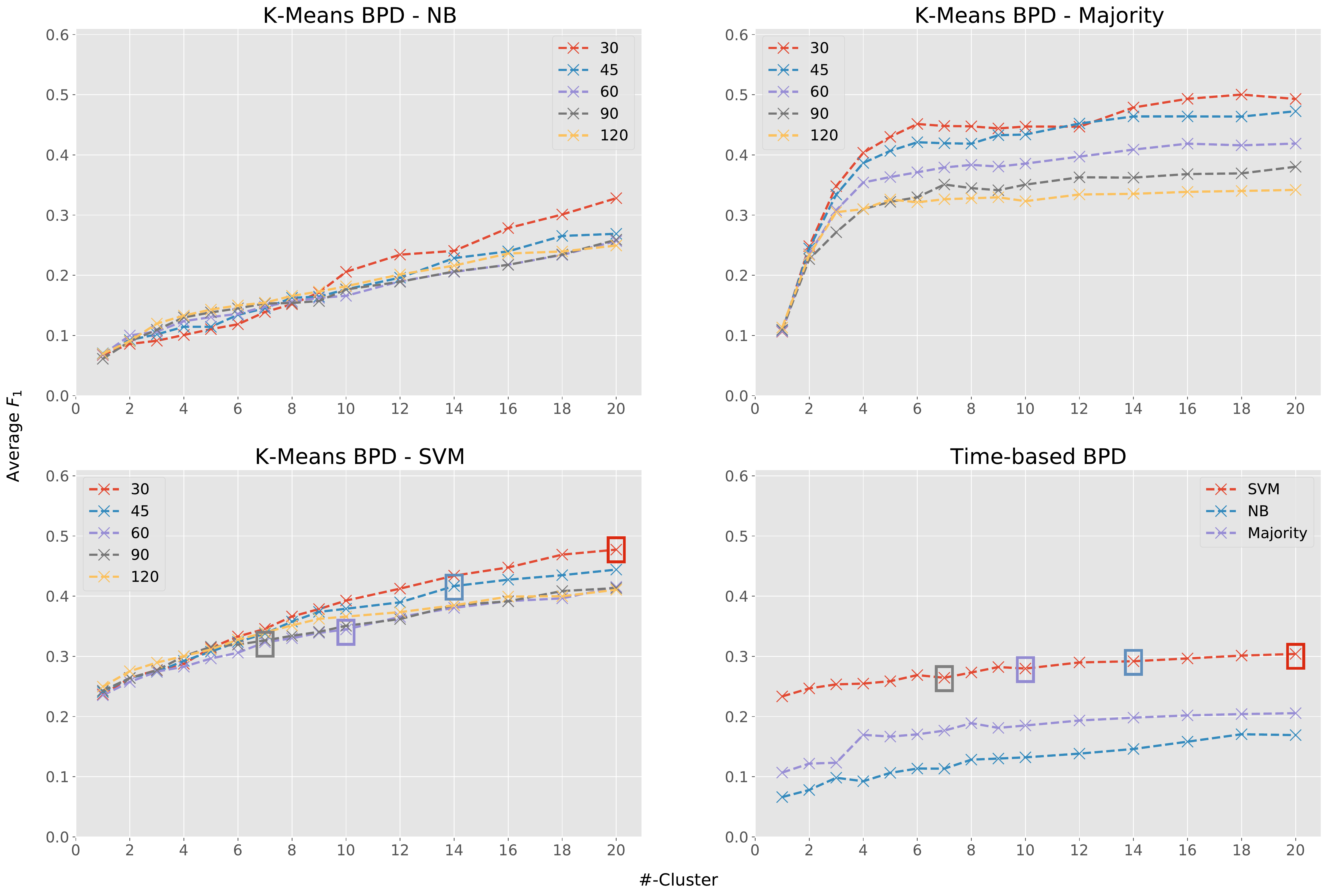}
  \caption{Averaged $F_1$ score for varying number of clusters and segment lengths. For time-based clustering, the segment length is implicitly given through the number of clusters. For k-means clustering, the segment lengths of the histograms are represented by the color. The segment length is in minutes. Rectangles denote similar BPD structures because of the same number of clusters and similar segment lengths.}
  \label{fig:f1s}
\end{figure*}

Figure \ref{fig:f1s} shows the average $F_1$ scores for each classifier and the strategies used for generating the BPDs. Firstly, we want to investigate only the k-means approach. The SVM shows an increase of $F_1$ score up to a number of about 10 clusters. From this point on, only smaller segment sizes benefit. For the majority classifier, the $F_1$ score no longer increases significantly above a number of 6 clusters. The naive Bayes classifier shows a nearly linear increase. Thus, for all classifiers, the $F_1$ measure increases with a larger number of clusters, as well as a decreasing segment length. The fact that the majority classifier benefits from this approach comes as no surprise. With an increasing number of clusters and finer segment lengths, the most frequent annotation occupies a higher portion of data within the cluster. Predicting only the most frequent annotation yields an $F_1$ score of around 0.1 whereas dividing the data by the BPD yields a score of around 0.5 at 20 BPDs and a segment length of 30 minutes. Note that some BPDs only describe one annotation and thus predicting can achieve 100\% accuracy by always outputting that one same class. According to this observation, the classification task should become simpler. Given such BPD with only one class, all classifiers used here can benefit and improve with knowledge of the BPD. However, the $F_1$ score of the majority classifier proposes that the knowledge of the most frequent annotation in the training data does not give us further advantages when we have more than 6 clusters. Yet the $F_1$ score of the naive Bayes classifier increases nearly linearly. Also, the SVM has a still increasing $F_1$ score at 20 BPDs, which underlines the fact that knowing the BPD simplifies the classification task by reducing the complexity of possible behaviors or even ruling them out. Note that the knowledge presented by the BPD converges with the BPD describing the exact underlying histogram, i.e. we have the same number of BPDs (clusters) as histograms.

We now investigate the time-based approach. The time-based approach shows an increase of $F_1$ score as the number of clusters rises. This can be observed for every classifier evaluated. Although the increase is not high, the knowledge of the clock time can help the classification. Thus dividing the day into segments and training a classifier for every segment each, increases the $F_1$ score. Comparing this approach with the k-means approach can be done by comparing the points where the number of BPDs and the segment length are approximately equal for both. E.g. splitting the day, where the subjects were recorded, into 20 segments yields a segment length of around 30 minutes and 20 BPDs. These points are denoted with rectangles in figure \ref{fig:f1s}. The plot yields a better performance of the k-means approach for all 4 points. The inferiority of the time-based approach can be explained through the high variances of annotations among the days (shown in figure \ref{fig:hist_violin}). The k-means approach, on the other hand, describes the histograms by their similarity and thus achieving less variance within the cluster. Having less variance means that the classifier has an easier task to solve.

\begin{figure}
  \centering
  \includegraphics[width=0.6\linewidth]{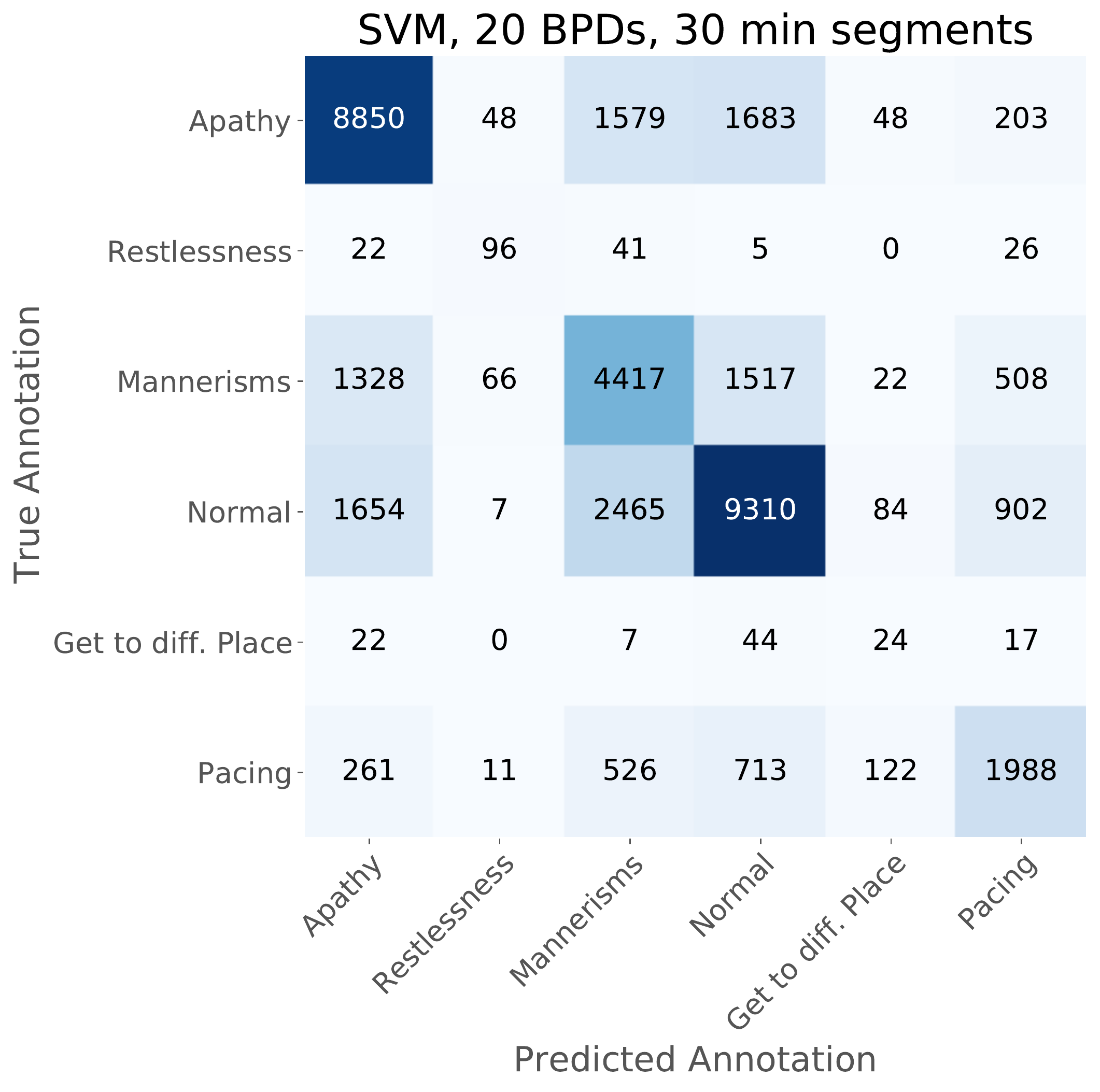}
  \caption{Confusion matrix of the predicted behaviors using 20 BPDs and thus 20 SVMs. A segment length of 30 minutes were used to calculate the histrogams.}
  \label{fig:confusion}
\end{figure} 

Overall, the $F_1$ score reaches a maximum of around 0.5. This score is produced by using 20 BPDs, a segment length of 30 minutes, and an SVM as a classifier. The same results can be achieved by using the majority method. The SVM outperforms the majority classifier for larger segment lengths. Figure \ref{fig:confusion} shows the confusion matrix of the SVMs for the classification with 20 BPDs and a segment length of 30 minutes. 
We can see that the classes mannerisms, apathy, and normal behavior represent the majority of the annotations. Furthermore, there is great uncertainty between these classes. This can be explained by the similarity of these behaviors in this scenario. Intuitively, normal behavior is similar to apathy, as dementia patients in the nursing home live more secluded. Mannerisms are difficult to recognize when, for example, they are performed with the non-dominant hand (sensor is worn at the dominant hand). Another uncertainty is between the annotations pacing, getting to a different place, and normal behavior. This can be explained by the fact that these three annotations can be understood as the activity of walking so a misclassification can occur between these classes. The behavior is also annotated in five-minute intervals, but the classification takes place in 30-second intervals. This means that the person may have walked at one point in time, but the interval was annotated differently. Another reason is the exclusive use of the motion sensor on the wrist.

\section{Future work}
The evaluation showed that the knowledge of the BPD can yield better classification and thus further investigations are of interest. There is the question of whether finding the BPD with k-means clustering is the best choice. Since similar histograms and thus categorical probability distributions are of interest, clustering methods with entropy-based distances, such as the Jensen-Shannon divergence, are of interest. Furthermore, Latent Dirichlet Allocation (LDA) \citep{lda} can be investigated as a method for determining behavioral dispositions. The LDA is a generative probabilistic model and is mainly used in applications to assign a topic to documents based on their words. In this scenario, a segment of the day can be regarded as a document and its words are the annotations in this segment so that the resulting topics describe the BPD. 

Currently, the BPD is only constructed by the annotations. A next step could be to determine the BPD based on the annotation as well as the sensor data. This would lead to BPDs that also hold information about the current motion data and its connection to the annotation. In this work, we created BPDs for a single person. As a next step, the evaluation of global BPDs across the patients is needed.

We also saw that the overall accuracy of the classification was low and thus looking into better options for classification is of interest. In this work, we only presented the approach of training a classifier for each BPD. However, this reduces the number of samples for the classifier. Another approach can be the feeding of the BPD label as input to the classifier. Also, classifiers that can make use of distributions over BPDs are of interest. Methods like mixed-effects neural networks \citep{menet} can model BPDs as another feature for activity recognition.

In reality, the BPD is not available and has to be found from sensor data. A good estimation of the BPD is therefore needed to improve the classification. With an increasing number of BPDs, the error of classification will also rise. Thus finding a good number of BPDs and a model to predict them is a crucial point for practical use. This can be done by using hierarchical models which first infer the BPD from the sensor data and then make use of the relations between behavior and BPD. Also, hidden Markov models can be of interest to exploit possible temporal properties of the BPDs.

\section{Related work}
To improve the classification of sensor data using additional information, firstly, the knowledge has to be modeled and perhaps has to be generated beforehand. Depending on the domain, the kind of knowledge can differ and thus arise different ways of incorporation into human activity recognition. Mostly the improvement of accuracy is the main purpose. However, the simplification of the recognition model can also be of interest. The authors of \citep{leehar} are aiming to create a model that still maintains the same level of accuracy under less training data than state-of-the-art methods. They use region connection calculus (RCC) \citep{rcc} to describe the spatial relations between objects and formalize the activities and their causal properties within the domain using the planning domain definition language (PDDL) \citep{pddl}. A probabilistic concurrent constraint automaton (PCCA) \citep{pcca} is generated from the PDDL model and reasons about the predicates from the RCC model. The RCC model calculates these predicates from high-level observations. The authors show that their approach can yield similar results in comparison to convolutional neural networks (CNN) but avoid the training data hungry nature of neural networks.
On the other hand, if enough training data and further information about the domain is available, the work of Xie et al. \citep{xie} aims to improve the accuracy of classification by incorporating prior knowledge in the training process of a neural network. They achieve this goal by modeling rules about the practicability of actions using linear temporal logics (LTL) \citep{ltl}. To overcome the problem of the multi-interpretable nature of LTL, the author convert them into deterministic finite automatons such that they can be embedded as hierarchical graph into the loss calculation of the neural network's training process and achieve learning consistent to the LTL rules and better accuracy at classification. 
The works by Yordanaova et al. \citep{yordanovacooking} and Rueda et al. \citep{rueda} use computational causal behavioral models (CCBM) \citep{ccbm} to model domain knowledge and reason about the current state of the environment. CCBM also uses PPDL-like rules to describe the domain-specific properties of the activities. Also, the relations between user and object as well as durations of activities can be modeled in CCBM. As an observation model, a decision tree (DT) in \citep{yordanovacooking} and a convolutional neural network in \citep{rueda} are used to weight the predicted state based on the sensor data. 
In contrast to these works, the prior knowledge does not need to be embedded directly in the model. L\"udtke et al. \citep{luedtkenn} demonstrate how providing external \emph{context information}  impacts classification performance. Specifically, they show that providing the current process step of structured activities can improve HAR performance, even when that information is noisy.
Even if no external information is available, the authors of \citep{inoue} made the observation that activities feature different quantities in times of a day. Thus segmenting the day and reasoning about the activities in one segment yields better accuracy than reasoning about the activities of the whole day.

\section{Conclusion}
In this paper, we proposed a data-driven method to construct behavioral predispositions and use them as informative prior to improve the performance of sensor-based human activity recognition. We presented a clustering-based approach that can determine time segments that display similar behaviors of a dementia-diagnosed person. We found such groups by first calculating the distribution of the annotations in the time segment and then clustering these distributions. The resulting clusters explain behavioral predispositions (BPDs). With complete knowledge of the current BPD, we demonstrated empirically that a substantial increase of the $F_1$ score can be achieved by using classifiers that were specifically trained for each BPD.
For practical usage, the investigation of predicting the BPDs from sensor data is of high interest, as the BPD is not given a priori and instead must be inferred. 

Furthermore, the presented approach is not only applicable to predicting the behavior of persons with dementia, but also to any continuous HAR task.
Whenever the prior for occurrence of activities varies over time, the method described in this paper offers a viable option to improve HAR performance.

\bibliographystyle{unsrtnat}
\bibliography{references}  






\end{document}